\documentclass[a4paper,fleqn]{cas-dc}

\usepackage[authoryear,longnamesfirst]{natbib}
\usepackage{algorithm}
\usepackage{algorithmic}
\usepackage{graphicx}
\usepackage{newfloat}
\usepackage{listings}
\usepackage{amsmath}     
\usepackage{amssymb}     
\usepackage{booktabs}
\usepackage{multirow}
\usepackage{subcaption}
\def\tsc#1{\csdef{#1}{\textsc{\lowercase{#1}}\xspace}}
\tsc{WGM}
\tsc{QE}
\tsc{EP}
\tsc{PMS}
\tsc{BEC}
\tsc{DE}

\begin{document}
\let\WriteBookmarks\relax
\def\floatpagepagefraction{1}
\def\textpagefraction{.001}

\shorttitle{Pattern Recognition Letters}

\shortauthors{CV Radhakrishnan et~al.}

\title [mode = title]{Mixture of Routers}                      
\cortext[1]{Corresponding author}
\author[1]{Jia-Chen Zhang}[orcid=0009-0004-5596-3952]
\ead{m325123603@sues.edu.cn}

\author[1]{Yu-Jie Xiong}[orcid=0000-0002-2769-022X]
\ead{xiong@sues.edu.cn}
\cormark[1]

\author[1]{Xi-He Qiu}
\ead{qiuxihe1993@gmail.com}

\author[1]{Chun-Ming Xia}
\cormark[1]
\ead{cmxia@sues.edu.cn}

\author[2]{Fei Dai}
\ead{fdai@fudan.edu.cn}

\author[1]{Zheng Zhou}
\ead{m320123332@sues.edu.cn}

\affiliation[1]{organization={School of Electronic and Electrical Engineering, Shanghai University of Engineering Science},
                addressline={\\333 Longteng Road, Putuo District}, 
                city={Shanghai},
                postcode={201620}, 
                state={Shanghai},
                country={China}}
                
\affiliation[2]{organization={Key Laboratory of Computational Neuroscience and Brain-Inspired Intelligence, Fudan University},
                addressline={\\220 Handan Road, Pudong New District}, 
                city={Shanghai},
                postcode={200433}, 
                state={Shanghai},
                country={China}}

\begin{abstract}
Supervised fine-tuning (SFT) is a resource-efficient approach for aligning large language models with human instructions. In particular, Low-Rank Adaptation (LoRA) has gained widespread attention due to its parameter efficiency. Recent studies suggest that combining LoRA with Mixture-of-Experts (MoE) significantly enhance fine-tuning performance. MoE enhances task accuracy by dynamically selecting the most suitable experts to adapt to diverse and complex datasets. Despite impressive results, recent studies reveal issues in the MoE routing mechanism, such as incorrect assignments and imbalanced expert allocation.
Previous researchers have alleviated this issue by controlling the number of experts or incorporating load balancing losses. However, single-router mechanisms still face challenges such as slow convergence when dealing with a large number of experts. Inspired by redundancy and fault tolerance theory, we are the first to propose the concept of MoE into the routing mechanism and propose an efficient fine-tuning method called Mixture of Routers (MoR). It employs multiple sub-routers for joint selection and uses a learnable main router to determine the weights of the sub-routers. Experimental results show that MoR outperforms baseline models on most tasks, with almost no increase in inference latency.
MoR can serve as a plug-and-play, parameter-efficient fine-tuning method for a wide range of applications. Our code is available here: \url{https://github.com/X-Lab-CN/MoR}.
\end{abstract}




\maketitle

\section{Introduction}

Large Language Models (LLMs) have gradually become the cornerstone of Natural Language Processing (NLP) \cite{bert, gpt-2,BUSTOCASTINEIRA202589}. As model parameters increase, LLMs demonstrate impressive emergent abilities and transfer learning capabilities \cite{emergent, palm, llama-adapter, mixtralexperts}. Therefore, Supervised Fine-Tuning (SFT) gradually became the mainstream solution for addressing various downstream tasks. However, the computational resources required for full fine-tuning were enormous \cite{full-finetuning}, and more and more research focused on parameter-efficient fine-tuning (PEFT) \cite{peft}. The main goal was to significantly reduce the resources required for fine-tuning. For example, P-tuning converted prompts into learnable embedding layers \cite{p-tuning}, while LoRA used a set of low-rank matrices to learn incremental updates \cite{lora}. Chain-of-LoRA \cite{10472574} reduced training resource consumption through LoRA networks. \cite{RASHID2023112} achieved unsupervised sentence simplification by fine-tuning BERT. Despite the high efficiency of PEFT methods, their fine-tuning performance often fell short of meeting the increasingly complex demands of downstream tasks.

Mixture-of-Experts (MoE) is designed to improve overall model performance by integrating the advantages of multiple expert networks \cite{mixtralexperts}. The core idea of this approach is that different expert networks can specialize in handling different subsets or features of the data, while a gating mechanism is responsible for determining which expert should process each input \cite{moe, sparsely-gated-moe}. The model typically includes multiple experts and a gating network. The task of the gating network is to evaluate the relevance of each expert based on the input data and dynamically allocate the input to the expert most suited to handle that data. This dynamic allocation mechanism enables MoE models to be more flexible and efficient, especially when dealing with large-scale and diverse datasets \cite{gshard, glam, widerinsteaddeeper, MoEbert}. Moreover, the MoE architecture offers higher computational efficiency and scalability. By parallelizing the computation across multiple experts, MoE can optimize resource usage, accelerating both training and inference speed. This is particularly important in scenarios that require processing massive datasets.

Recent research has shown that combining PEFT and MoE enables the model to leverage the advantages of both approaches \cite{zadouri2024pushing}. LoRAMoE \cite{dou2024loraMoE} introduced a plug-in version of MoE that learned multiple sets of low-rank matrices as experts and used a softmax layer as a router to compute each expert’s contribution for a given input. During training, LoRAMoE kept the pre-trained weights frozen and trained only the experts and the router. MoLA \cite{gao-etal-2025-mola} further investigated expert assignment across layers, assigning fewer experts to lower layers and increasing the number of experts with depth. On the other hand, recent studies have identified issues in MoE routing mechanisms, such as incorrect expert assignments and imbalanced expert utilization \cite{shazeer2017,10.5555/3586589.3586709}.

In system design, the principles of Redundancy and Fault Tolerance Theory emphasize the importance of using multiple components to enhance reliability and robustness. By introducing redundancy, systems can mitigate the impact of individual component failures and improve overall performance. Inspired by this theory, we propose a new parameter-efficient MoE method to address the aforementioned issues. Our approach employs multiple sub-routers for joint decision-making, where each sub-router contributes to the final decision, thereby reducing the risk of errors from any single sub-router. A main router is used to assign weights to the sub-routers, ensuring that the most reliable decisions are prioritized. Similar to MoE adjusting the number of experts. By adjusting the number of sub-routers, MoR can flexibly adapt to tasks of varying complexity. Ultimately, the weighted cooperation of the sub-routers determines the scores for each expert, and the top-k experts with the highest combined scores are selected for the final inference. Additionally, MoR further balances expert load by introducing load balancing loss.
We conducted experiments on six benchmarks, including NLP and Commonsense Reasoning (CR) tasks, to demonstrate the effectiveness of MoR. Our main contributions are as follows:
\begin{itemize}
\item[$\bullet$] We propose a new fine-tuning method called MoR, which selects expert models through multiple sub-routers and uses a main router to determine the selection of sub-routers. MoR can replace the router layer in MoE-style models, thereby making it a plug-and-play and parameter-efficient solution.
\item[$\bullet$] The impact of the number of sub-routers on training time and performance is analyzed to facilitate determining the optimal number of sub-routers.
\item[$\bullet$] Numerous experiments are conducted to validate the effectiveness of MoR. We compare it against benchmarks across six different tasks, and MoR outperforms in most of them.
\end{itemize}

\section{Related work}
When performing SFT tasks, full fine-tuning not only requires substantial computational and storage resources but can also lead to catastrophic forgetting. In contrast, PEFT \cite{peft} achieves similar results to full fine-tuning by freezing most of the model parameters and training only a small subset of them. Low-Rank Approximation \cite{lora} is a popular and efficient fine-tuning method for LLMs, dubbed as LoRA. It utilizes low-rank approximation theory to effectively adjust the model's behavior with smaller parameter increments. 
Building upon this low-rank adaptation framework, PiSSA \cite{NEURIPS2024_db36f4d6} initialized the adapter matrix using the singular values of the original weight matrix and placed the remaining components into a residual matrix. DCFT \cite{zhang-etal-2025-parameter} significantly reduced the number of trainable parameters by completing the incremental update matrix with deconvolution. Bpc-lw \cite{ZHANG202581} fine-tuned the BERT model via LoRA to achieve few-shot HTC tasks. LoRAC \cite{LING2026112086} proposed an orthogonal LoRA combination based on QR decomposition, which can preserve previously learned knowledge while integrating new knowledge into continual learning.

Mixture-of-Experts (MoE) architectures enhance model capacity and performance by leveraging multiple specialized expert subnetworks, each trained to handle distinct data patterns or input regions \cite{mixtralexperts}. A learnable gating network dynamically routes each input to the most relevant expert(s), enabling conditional computation and improved adaptability \cite{sparsely-gated-moe}. This design not only increases model expressiveness but also improves computational efficiency through sparse activation—only a subset of experts is activated per input.

Although LoRA significantly reduces the number of parameters, its impact on SFT performance is noticeable. \cite{zadouri2024pushing} introduce a novel parameter-efficient MoE framework, MoV and MoLORA, which achieves comparable performance to full fine-tuning with significantly fewer parameters. \cite{huang2024lorahub} propose LoRAHub, which achieves the cross-task generalization capability of LoRA by integrating LoRA modules trained on different tasks through a simple framework. MoLA \cite{gao-etal-2025-mola} experimentally demonstrates that more LoRA experts in higher layers can significantly improve the performance of Transformer-based models.

These methods effectively improved the performance of supervised fine-tuning (SFT), but they also introduced issues such as routing errors and imbalanced expert allocation \cite{shazeer2017,10.5555/3586589.3586709}. To mitigate the random routing phenomenon in MoE, MoELoRA \cite{Luo2024MoELoRACL} encouraged experts to learn distinct features through contrastive learning, thereby improving model performance. LoRAMoE \cite{dou2024loraMoE} integrated multiple LoRA experts via a router and further alleviated unbalanced expert utilization through a Localized Balancing Constraint. DeepseekV3 \cite{DeepSeekAI2024DeepSeekV3TR} addressed load imbalance using Auxiliary-Loss-Free Load Balancing together with Complementary Sequence-Wise Auxiliary Loss. While these approaches provided valuable insights into routing errors and expert imbalance, MoR offers a more flexible and plug-and-play strategy to tackle these challenges.

\begin{figure*}
\centerline{\includegraphics[width=\textwidth]{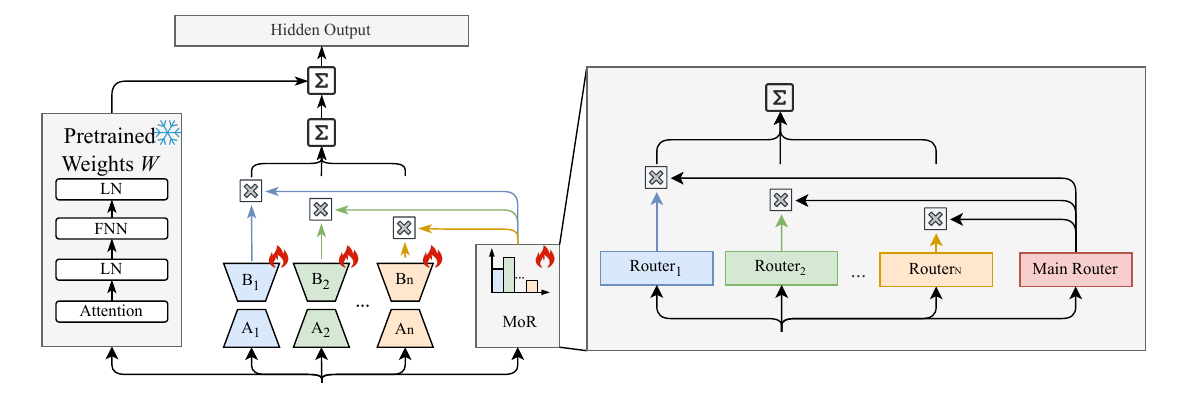}}
  \caption{Here is a schematic illustration of MoR. On the left is a schematic diagram of the integration of LoRA and MoE, incorporating LoRA as an expert into the MoE model. On the right is the MoR plugin we propose, which can flexibly replace the router layer in MoE-style models.}
  \label{photomain}
\end{figure*}

\section{Method}
\label{method}
In this section, we elaborate on the methodological details of MoR. It is a MoE-style plug-in module that employs multi-route collaborative expert allocation, as illustrated in Figure~\ref{photomain}.

\subsection{Motivation}
In this section, we explore why we introduce multiple routers to make joint decisions and explain their necessity and potential advantages based on the theory of redundancy and fault tolerance.

In traditional MoE-style architectures, a single router is responsible for assigning inputs to the most suitable expert network. However, this design has a significant limitation: the single router may make incorrect allocation decisions due to noise, overfitting, or insufficient training, which can lead to a decline in model performance. This vulnerability is particularly pronounced in high-dimensional and complex tasks. The theory of redundancy and fault tolerance provides important inspiration for addressing this issue. The theory suggests that by introducing redundant components and designing appropriate fault-tolerant mechanisms, the reliability and robustness of a system can be significantly improved. Specifically, when one component in the system fails, other components can collaborate to take over its tasks, thereby preventing overall system failure. Inspired by this, we introduce multiple routers in the MoE architecture to make joint decisions.
Different routers can form complementary decision-making capabilities through varied initializations, training data, or structural designs. This diversity not only strengthens the system's robustness but also improves the model's adaptability when facing complex tasks.
\subsection{Mixture of Routers}
The left of Figure \ref{photomain} illustrates the forward process of the standard LoRAMoE architecture. MoE assigns different inputs to different experts via a router module. This means that although adding more experts increases the total number of model parameters, only a small number of experts are involved in the computation during both training and inference. This allows the model parameters to scale with the same computational cost. The router is a trainable gating function that normalizes the distribution of expert weights using the Softmax function. The final output of the MoE layer is the weighted sum of the outputs from the experts:
\begin{equation} 
\label{eq:1} 
F_{i}(x) = \frac{\text{Topk}(\text{Softmax}(W_r \cdot x), k)_i}{\sum_{j=1}^{k} \text{Topk}(\text{Softmax}(W_r \cdot x), k)_j},
\end{equation} 
\begin{equation} 
\label{eq:2} 
W = W^{0}+\sum_{i=1}^{k}F_{i}(x) \cdot E_i(x),
\end{equation} 
where $E_i(x)$ denotes the output of the $i$-th expert. Topk retains the $k$ highest weight distributions P from the Softmax output, and the remaining weights are set to 0. The retained $k$ weights are then re-normalized to ensure the sum of the weights equals 1. In addition, load balancing loss is commonly used in MoE to promote balanced expert selection.
To ensure that the original knowledge space of LLMs is not compromised, LoRA is adopted to reduce the occurrence of catastrophic forgetting. By training multiple pairs of low-rank matrices $\{A\}_{i=1}^{N}$ and $\{B\}_{i=1}^{N}$ as experts. The input is then assigned to different experts through a learnable routing module. Our forward equation is shown as follows:
\begin{equation} 
\label{eq:3} 
W = W^{0}+\sum_{i=1}^{k}F_{i}(x) \cdot B_iA_i(x).
\end{equation} 

The right of Figure \ref{photomain} illustrates the MoR plugin. In previous MoE models, experts were selected based on the results of main router $W_R$, typically choosing the top one or two experts determined by the top-k criterion:
\begin{equation} 
\label{eq:4} 
R^{i} = \text{Softmax}(W_R \cdot x)_i,
\end{equation} 
while it improves the coordination ability of expert models in handling complex tasks, it still faces the issues of incorrect and uneven expert allocation. To address the issues, we propose a fine-grained expert control method with a multi-router mechanism, termed MoR, which can be inserted into all MoE-based models by replacing the router component of the original model. It consists of a main router $W_R$ and multiple sub-routers $W_{r}$:
\begin{equation} 
\label{eq:5} 
F_{i}(x)' = \sum_{i=1}^{N}\frac{R^i}{\sum_{j=1}^{k}R^j} \cdot r^i,
\end{equation} 
where the main router assigns weights to each of the sub-routers. The final expert routing weight $R'$ is a weighted sum of the sub-routers rather than relying on the result of a single router. Load balancing loss is employed to promote balanced router selection. It is used to ensure that each expert is utilized reasonably when handling tasks, avoiding situations where some experts are overloaded while others remain idle. By introducing load balancing loss, the model can dynamically adjust the routing mechanism during training, enabling a more even distribution of input samples among different experts. The calculation formula of $L_{\text{balance}}$ is as follows:
\begin{equation} 
\label{eq:6} 
L_{\text{balance}} = \sum_{i=1}^{N_{\text{layers}}} \left( N_i \cdot \sum_{j=1}^{N_i} t_{j,i} \cdot R_{j,i} \right),
\end{equation} 
where $ N_i $ is the total number of experts in the $ i $-th layer. $ t_{j,i} $ is the proportion of tokens assigned to the $ j $-th expert in the $ i $-th layer. $ r_{j,i} $ represents the average routing probability of the $ j $-th expert in the $ i $-th layer.


\section{Experiments}
\subsection{Experimental settings}
\label{setting}
Similar to \cite{gao-etal-2025-mola}, we design two experimental setups to evaluate the performance of the MoR, including direct fine-tuning and transfer fine-tuning. Direct fine-tuning refers to fine-tuning the model directly on downstream tasks, while transfer fine-tuning involves first performing preliminary fine-tuning on an instruction-tuning dataset (OpenOrca), followed by secondary fine-tuning on the downstream task. The direct fine-tuning experiments use LLaMA-7B-hf \cite{touvron2023llama}, LLaMA-2-7B-hf \cite{llama2}, and Mistral-7B-v0.1 \cite{jiang2023mistral7b} as the base models.
To ensure a fair comparison, LoRAMoE allocates 5 experts per layer, while MoLA adopts a similar progressive expert configuration of 2-4-6-8 as mentioned in its paper. The total number of experts was held constant across all variants. In both settings, we conduct a grid search on the number of training epochs, considering 10, 15, and 20 epochs for fine-tuning on the downstream task. The highest value from the three experiments is taken as the experimental result. We use AdamW \cite{adamw} as the optimizer. We applied LoRA to four weight matrices in the self-attention module $(W_q, W_k, W_v, W_o)$ and to three weight matrices in the MLP module $(W_{gate}, W_{down}, W_{up})$. The MoR results in the main experiments are all based on tests with dual sub-routers. All experiments are conducted on a single NVIDIA A100-80G GPU.

\begin{table*}
\centering
\tabcolsep=0.25cm
\renewcommand\arraystretch{1.0}
\caption{Direct Fine-Tuning Performance comparison of different models on various tasks. We report the accuracy of all the tasks. Higher is better for all metrics. The best results are denoted in \textbf{bold}.}
\begin{tabular}{l|cccccc|l}
\toprule
\textbf{Models} & \textbf{SciQA} & \textbf{ComQA} & \textbf{OpenQA} & \textbf{MRPC} & \textbf{CoLA} & \textbf{RTE} & \textbf{Avg.}\\
\midrule
Llama-7B-hf\\
\midrule
LoRA &89.12 &73.96 &75.80 &82.43&84.18&83.03&81.42\\
LoRAMoE &90.28&75.10&76.00&82.43&84.18&83.03&81.84\\
\textbf{LoRAMoE + MoR} &\textcolor{red}{(+0.55)}&\textcolor{red}{(+0.58)}&\textcolor{red}{(+0.40)}&\textcolor{red}{(+2.15)}&\textcolor{red}{(+1.63)}&\textcolor{red}{(+0.72)}&82.84 \textcolor{red}{(+1.00)}\\
MoLA &90.10&75.42&78.60&83.36& 84.64&84.43&82.76 \\
\textbf{MoLA + MoR} &\textcolor{red}{(+0.91)}&\textcolor{red}{(+1.24)}&\textcolor{blue}{(-0.20)}&\textcolor{red}{(+0.99)}&\textcolor{red}{(+0.88)}&\textcolor{red}{(+0.41)}&83.46 \textcolor{red}{(+0.70)}\\
\midrule
Llama-2-7B-hf\\
\midrule
LoRA & 91.01 & 75.51 & 77.00& 83.13 & 86.29 & 85.92 & 83.14\\
LoRAMoE & 92.04 & 78.13 & 80.00 & 84.23 & 86.28 & 85.20 &84.31\\
\textbf{LoRAMoE + MoR} &\textcolor{red}{(+0.86)} &\textcolor{red}{(+0.41)} &\textcolor{red}{(+1.00)} &\textcolor{red}{(+0.62)} &\textcolor{red}{(+0.11)} &\textcolor{red}{(+3.25)} &85.34 \textcolor{red}{(+1.04)}\\
MoLA & 92.36 &  78.95 & 79.60 & 83.48 &  86.87 & 86.28 &84.59\\
\textbf{MoLA + MoR} & \textcolor{red}{(+0.72)} & \textcolor{red}{(+0.25)} &  \textcolor{red}{(+2.40)} &  \textcolor{red}{(+0.46)} & \textcolor{red}{(+0.00)}&  \textcolor{red}{(+2.17)} &85.57 \textcolor{red}{(+0.99)} \\
\midrule
Mistral-7B-v0.1\\
\midrule
LoRA &91.97&80.86&87.20&86.83&87.92&89.53&87.39\\
LoRAMoE &94.46&81.90&88.00&85.73&87.34&88.44&87.65\\
\textbf{LoRAMoE + MoR} &\textcolor{red}{(+0.56)}& \textcolor{red}{(+1.47)}&\textcolor{red}{(+0.00)} &\textcolor{red}{(+1.46)} &\textcolor{blue}{(-0.28)} &\textcolor{red}{(+1.09)}&88.36 \textcolor{red}{(+0.71)}\\
MoLA &94.91&82.96&88.00&86.95&87.24&88.80&88.32 \\
\textbf{MoLA + MoR} &\textcolor{red}{(+0.23)} &\textcolor{red}{(+0.00)} &\textcolor{red}{(+0.20)} &\textcolor{red}{(+0.64)} &\textcolor{red}{(+0.10)} &\textcolor{red}{(+2.53)}& 88.76 \textcolor{red}{(+0.44)} \\
\bottomrule
\end{tabular}
\label{main}
\end{table*}
\subsection{Main results}
\subsubsection{Direct fine-tuning} 
We first compare the results of direct fine-tuning between MoR and baseline models on three commonly used models (Llama-7B-hf, Llama-2-7B-hf and Mistral-7B-v0.1). The accuracy results of MoR and other baseline models are shown in Table \ref{main}. We use the same hyperparameters for all methods. The results indicate that after inserting the MoR module, there is a significant improvement across most tasks for all models. Taking the Llama-2-7B-hf model as an example. After integrating the MoR module, MoELoRA and MoLA show increased accuracy rates in the OpenbookQA task by 1.4\% and 2.4\%, respectively, and in the RTE task by 2.17\% and 1.58\%, respectively. On average, the insertion of the MoR module has led to performance improvements of 1.03\% and 0.98\%, respectively. Consistent improvements are observed on the other two models, showing average performance gains of 0.71\% and 0.44\% after integrating MoR. These results demonstrate that MoR can effectively enhance the performance of MoE-style models.

\subsubsection{Transfer fine-tuning}In the context of transfer fine-tuning, we first fine-tune LLAMA-2-7B-hf on an instructional tuning dataset using each PEFT (Parameter-Efficient Fine-Tuning) method. Subsequently, we further fine-tune the model on all downstream tasks. This instructional tuning process serves as an effective means to evaluate the transfer learning capabilities of each PEFT method. To ensure a focused comparison, we limit our analysis to methods based on LoRA, as these have demonstrated superior transfer learning performance compared to approaches reliant on prompt tuning. As shown in Table \ref{main1}, MoR continues to outperform other methods across most benchmark tests. Notably, in the CommonsenseQA and OpenbookQA tasks, MoLA+MoR achieves significant improvements over MoLA, with performance gains of 2.9\% and 0.82\%, respectively. On average, the incorporation of MoR leads to a notable 1.16\% improvement in overall performance. These experimental results underscore the value of MoR as a plug-and-play enhancement for MoE-style models. By integrating MoR, the overall performance of MoE models is significantly boosted, while also showcasing its robust transfer learning capabilities. This makes MoR a highly versatile and effective solution for a wide range of applications.
\begin{table*}
\centering
\tabcolsep=0.25cm
\renewcommand\arraystretch{1.0}
\caption{The transfer learning capabilities of different models across various tasks. We use OpenOrca as the dataset for the first fine-tuning.}
\begin{tabular}{l|cccccc|l}
\toprule
\textbf{Models} & \textbf{SciQA} & \textbf{ComQA} & \textbf{OpenQA} & \textbf{MRPC} & \textbf{CoLA} & \textbf{RTE} & \textbf{Avg.} \\
\midrule
LoRA & 91.01 & 74.61 & 76.6 & 84.41 & 84.95 & 84.48 & 83.01 \\
LoRAMoE&91.41&75.92&77.6&84.93&85.81&86.64&83.72\\
\textbf{LoRAMoE + MoR}&\textcolor{red}{(+0.95)}&\textcolor{red}{(+1.23)}&\textcolor{red}{(+0.80)}&\textcolor{red}{(+0.00)}&\textcolor{red}{(+0.48)}&\textcolor{red}{(+1.09)}&84.48 \textcolor{red}{(+0.76)}\\
MoLA & 92.94 & 77.97 & 78.7 & 84.52 & 86.64 & 86.19 &84.54\\
\textbf{MoLA + MoR} & \textcolor{red}{(+0.27)} & \textcolor{red}{(+0.82)} & \textcolor{red}{(+2.9)} & \textcolor{red}{(+0.87)} & \textcolor{red}{(+0.23)}& \textcolor{red}{(+2.26)} &85.70 \textcolor{red}{(+1.16)}\\
\bottomrule
\end{tabular}
\label{main1}
\end{table*}

\begin{table}
\centering
\tabcolsep=0.1cm
\renewcommand\arraystretch{1.0}
\caption{The results of MoR with different router numbers were tested on 6 different tasks, and we report the accuracy and time efficiency. $r$ represents the number of routers, $h$ and $m$ represent hours and minutes, respectively.}
\begin{tabular}{c|c|cccc}
\toprule
 {\bf Method} & \textbf{Metric} & \textbf{ComQA} & \textbf{OpenQA}& \textbf{CoLA} & \textbf{RTE} \\
\midrule 
\multirow{3}{*}{\bf MoLA} & {Acc(\%)}  &  78.95 & 79.6 & 83.48 &  86.77 
\\
& {Time(train)} & {13.20h}& {9.67h}& {7.52h}& {15.80h}
\\
& {Time(test)}  & {6.57m}& {2.53m}& {8.07m}& {1.50m}
\\
\midrule 
\multirow{2}{*}{\bf+ MoR}& {Acc(\%)} & \textbf{79.20} &  \textbf{82.0} &  83.94 & \textbf{86.87}
\\
 &  {Time(train)}  & {19.78h}& {10.03h}& {7.92h}& {17.00h}
\\
{\bf $(r=2)$} & {Time(test)}  & {7.14m}& {2.55m}& {8.32m}& {1.53m}
\\
\midrule
\multirow{2}{*}{\bf + MoR}& {Acc(\%)}  & {78.05}& {81.2}& {\bf84.81}& {86.10}
\\
 & {Time(train)} & {19.99h}& {10.27h}& {7.98h}& {17.15h}
\\
{\bf $(r=3)$} &{Time(test)} & {7.29m}& {3.05m}& {8.40m}& {1.55m}
\\
\midrule
\multirow{2}{*}{\bf+ MoR}& {Acc(\%)}& {76.82} & {80.0}& {84.00}& {86.10}
\\
& {Time(train)} & {20.31h}& {10.57h}& {8.02h}& {17.31h}
\\
{\bf $(r=4)$} &{Time(test)}  & {7.34m}& {3.11m}& {8.59m}& {2.05m}
\\
\bottomrule
\end{tabular}
\label{tab:1}
\end{table}
\subsection{Analysis}
\subsubsection{Router number analysis}
\label{number}
Similar to MoE, MoR can also better adapt to different downstream tasks by adjusting the number of routers. In this section, we present the results and training times of MoR models with different numbers of routers across six distinct tasks, as shown in Table~\mbox{\ref{tab:1}}.
The analysis shows that when using two routers ($r=2$) for joint allocation, the model achieves the best overall results, with an average improvement of 0.98\%. When using three routers ($r=3$) for joint allocation, the model achieves average suboptimal results; however, it is noteworthy that the model produces the best outcomes on the ScienceQA and MRPC tasks. Upon further increasing the number of routers to four ($r=4$), there is a decline in model performance. 

Further analysis indicates that ScienceQA and MRPC are the most complex tasks among similar types of three tasks, thus requiring more joint allocation experts through routing. In contrast, simpler tasks are prone to overfitting due to excessively high model complexity when using multiple routers. Moreover, unlike MoE, which requires dynamic selection among all experts for each input, MoR only needs to coordinate and make decisions among a small number of routers. Specifically, our results show that using a number of routers equivalent to approximately one-fourth the number of experts is sufficient to achieve near-optimal performance. This significantly reduces the complexity of routing decisions, as the main router in MoR only needs to assign weights among a few sub-routers. Consequently, MoR largely avoids the issue of imbalanced routing load that is commonly observed in MoE systems. 
In summary, the model performs best when using two routers ($r=2$) for joint allocation, achieving improved performance while maintaining high training efficiency.

\subsubsection{Efficiency analysis}
In this section, we conduct an analysis of the efficiency of MoR, with the relevant results presented in Table \ref{tab:1}. The table provides a detailed breakdown of how different numbers of sub-routers impact training and inference efficiency. Experimental data shows that our best-performing model, MoLA+MoR $(r=2)$, introduces relatively low latency during both training and inference. Specifically, training time increases by an average of 4.41\%, while inference time rises by an average of 2.58\%. Although these additional costs are not negligible, they are entirely acceptable when compared to the significant performance improvements achieved by the model. Furthermore, the experimental results further validate the rationality of the model design, demonstrating that MoR can achieve a good balance between efficiency and performance when the number of sub-routers is moderate.


\subsubsection{Expert allocation analysis}
\begin{figure}[t]
\centerline{\includegraphics[width=1.0\columnwidth]{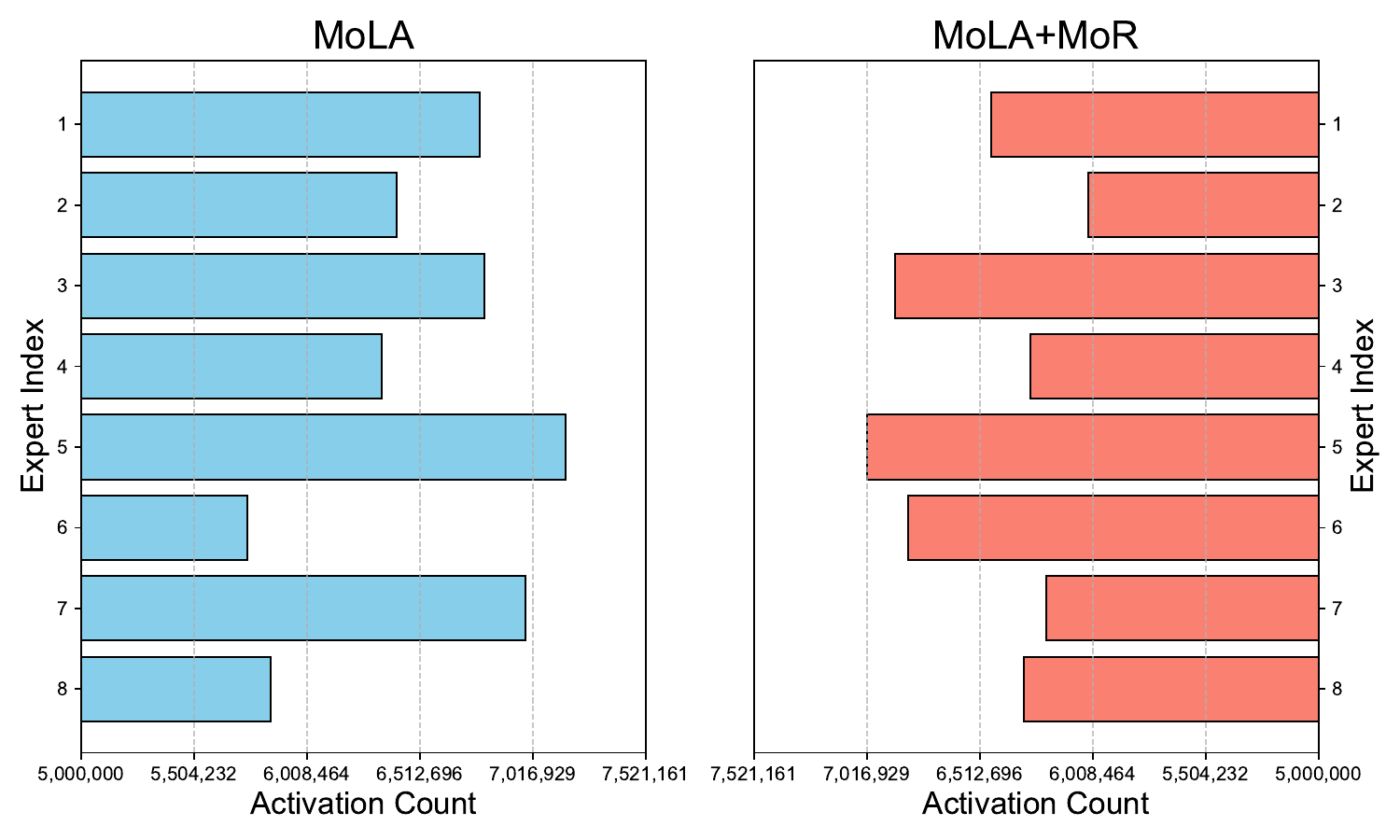}}
  \caption{Visualizing the effect of incorporating the MoR plugin into the model through expert activation patterns with 8 experts.}
  \label{expert8}
\end{figure}

In this section, we evaluate how MoR affects expert allocation by recording the number of expert activations. The changes in expert allocation after inserting MoR are shown in Figure \ref{expert8}. Before inserting MoR, there was a significant difference between the most frequently activated experts (5, 7) and the least frequently activated experts (6, 8), reflecting an imbalance in workload distribution within the original MoE architecture. Similar activation patterns can also be observed under other configurations of the number of experts. However, after inserting MoR, the distribution of expert activations becomes more balanced, and the difference between the most and least activated experts is significantly reduced.
Specifically, the MoR plugin introduces a multi-routing mechanism, allowing the model to consider more information when selecting experts. This avoids over-reliance on a single expert or neglecting certain experts.
This improvement in load balancing not only enhances training efficiency but also reduces potential biases caused by uneven expert allocation.

\section{Conclusion}
In this work, we identify a critical yet underexplored limitation in current MoE based parameter-efficient fine-tuning (PEFT) approaches: the fragility of single-router mechanisms, which leads to misrouted expert assignments and severe load imbalance—especially as model scale and expert count grow. To address this, we propose Mixture of Routers (MoR), a novel plug-and-play PEFT framework that rethinks routing not as a monolithic decision, but as a collaborative, redundant process inspired by fault-tolerant system design.

MoR introduces multiple learnable sub-routers that jointly evaluate expert suitability, while a lightweight main router dynamically arbitrates their contributions. This multi-path routing strategy significantly enhances assignment accuracy and expert utilization balance, without incurring additional inference latency—a crucial requirement for real-world deployment. Extensive experiments across NLP and commonsense reasoning benchmarks demonstrate that MoR consistently outperforms strong MoE-LoRA baselines, achieving an average performance gain of 1\% under both standard and instruction-based fine-tuning paradigms.
\\ \\
\textbf{Acknowledgments}
\\
\par This work was supported in part by the Science and Technology Commission of Shanghai Municipality under Grant (21DZ2203100), in part by the National Natural Science Foundation of China under Grant (62006150).
\\ \\
\textbf{Data availability}
\\
\par The data used in this study are publicly available.

\printcredits

\bibliographystyle{cas-model2-names}

\bibliography{cas-refs}

\end{document}